\documentclass[runningheads]{llncs}

\usepackage{eccv}

\usepackage{eccvabbrv}

\usepackage{graphicx}
\usepackage{booktabs}

\usepackage[accsupp]{axessibility}  

\usepackage{multirow}
\usepackage{booktabs}
\usepackage{array}
\usepackage{xcolor}

\usepackage{hyperref}

\usepackage{orcidlink}

\begin{document}

\title{RenderMem: Rendering as Spatial Memory Retrieval} 

\titlerunning{RenderMem}

\author{JooHyun Park\inst{1}\orcidlink{0009-0009-5891-7405} \and
HyeongYeop Kang\inst{1}\orcidlink{0000-0001-5292-4342}}

\authorrunning{J. Park et al.}

\institute{Korea University, South Korea}

\maketitle

\begin{abstract} 
Embodied reasoning is inherently viewpoint-dependent: what is visible, occluded, or reachable depends critically on where the agent stands. However, existing spatial memory systems for embodied agents typically store either multi-view observations or object-centric abstractions, making it difficult to perform reasoning with explicit geometric grounding.
We introduce \textbf{RenderMem}, a spatial memory framework that treats rendering as the interface between 3D world representations and spatial reasoning. Instead of storing fixed observations, RenderMem maintains a 3D scene representation and generates query-conditioned visual evidence by rendering the scene from viewpoints implied by the query. This enables embodied agents to reason directly about line-of-sight, visibility, and occlusion from arbitrary perspectives.
RenderMem is fully compatible with existing vision–language models and requires no modification to standard architectures. Experiments in the AI2-THOR environment show consistent improvements on viewpoint-dependent visibility and occlusion queries over prior memory baselines.
\end{abstract}

\begin{figure}[t]
    \centering
    \includegraphics[width=\textwidth]{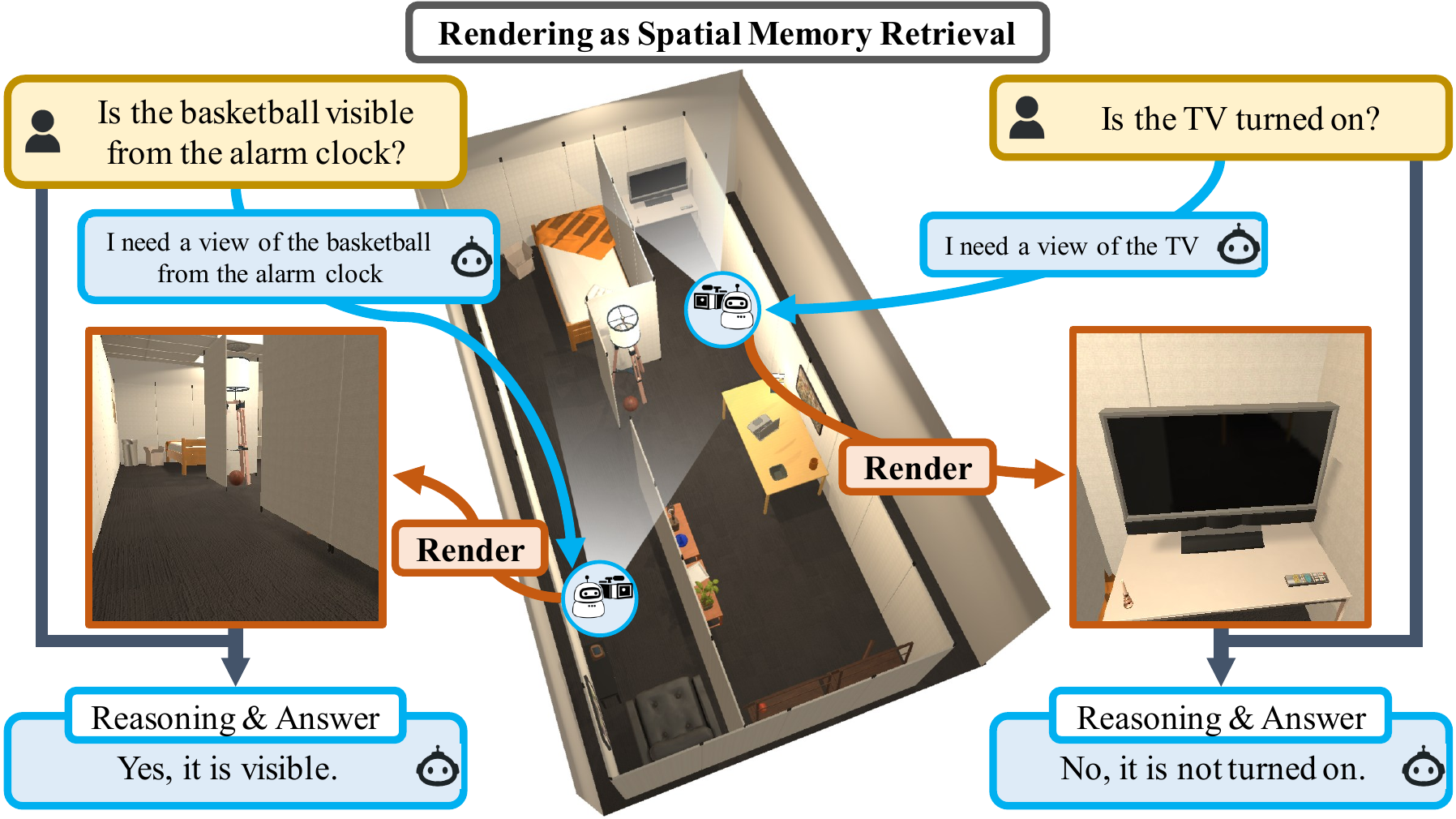}
    \caption{RenderMem retrieves spatial evidence from a stored 3D scene by rendering query-conditioned views, which serve as visual memory readouts for vision--language reasoning about visibility and object state.}
    \label{fig:teaser}
\end{figure}

\section{Introduction}
\label{sec:intro}
Embodied reasoning is fundamentally viewpoint-dependent. In the physical world, perception is spatially situated: what an agent can observe depends on where it stands, and the same scene can yield entirely different evidence under different viewpoints. Because action decisions are grounded in what is visible from a specific location, visibility directly constrains movement, manipulation, and task execution. Viewpoint-conditioned reasoning is therefore intrinsic to embodied intelligence.

Consider a robot responsible for inspecting safety conditions in a building. To ensure emergency equipment can be located quickly, it may need to reason: ``From the main corridor, is the fire extinguisher visible, or is it occluded by the cabinet?'' Answering such questions requires interpreting the corridor as a reference viewpoint and evaluating line-of-sight between objects. This is not merely object recognition—it is geometric reasoning over spatial configuration.

Despite recent progress, existing spatial memory architectures struggle with such viewpoint-conditioned queries. Current approaches largely fall into three paradigms, each sharing a structural limitation.

View-based memory~\cite{yang20253d, wang2026explore} stores observations from discrete viewpoints and retrieves images for downstream reasoning. While effective for recognizing visible attributes, it is constrained by previously captured views and cannot reliably answer queries requiring novel or object-centric viewpoints.

Object-centric memory~\cite{armeni20193d, gu2024conceptgraphs} represents scenes as collections of objects and relations. Although compact and suitable for relational reasoning, it typically lacks explicit modeling of camera pose and line-of-sight geometry, making visibility reasoning difficult.

3D scene representations~\cite{zhang2023clip, hu20253dllm} encode richer geometric structure through meshes, volumetric maps, or neural fields. Yet their high-dimensional nature makes direct integration with language models challenging, often weakening the link between geometry and reasoning.

These limitations point to a deeper issue: existing memory systems retrieve stored observations or abstract relations, but do not actively generate viewpoint-specific visual evidence required by a query. We argue that resolving viewpoint-dependent reasoning requires rethinking memory access itself.

In this work, we introduce \textbf{RenderMem}, a rendering-based spatial memory framework built on a simple but powerful principle: \emph{rendering is the read operation of 3D memory}. 
RenderMem maintains a persistent 3D scene representation and, when faced with a viewpoint-conditioned query, performs \emph{query-conditioned rendering} to synthesize exactly the visual evidence needed for reasoning, as illustrated in \cref{fig:teaser}. 
Instead of forcing language models to interpret raw 3D geometry, RenderMem translates geometry into images aligned with the query's specified viewpoint, which are then processed by standard vision--language models. 
By turning rendering into a first-class memory primitive, RenderMem enables explicit, geometrically grounded visibility and occlusion reasoning that prior memory paradigms cannot reliably support.

Importantly, this design is not limited to static scenes. 
Because memory is maintained as an updatable 3D representation, any modification to object state or geometry is immediately reflected in subsequent query-conditioned renderings. 
Thus, RenderMem naturally extends to dynamic environments, serving as a spatial memory that evolves consistently with scene changes rather than requiring explicit rewriting of stored observations.

Our contributions are as follows:
\begin{itemize}
\item We identify viewpoint-dependent visibility and occlusion reasoning as a fundamental and under-addressed bottleneck in embodied spatial memory.
\item We propose RenderMem, introducing the abstraction of rendering as a query-conditioned memory read operation, enabling geometrically grounded reasoning without modifying existing vision–language architectures.
\item We develop query-conditioned viewpoint synthesis strategies that explicitly support visibility and occlusion reasoning, bridging the gap between 3D geometry and language-based inference.
\item Experiments show that RenderMem achieves superior performance over view-based and object-centric memory baselines while remaining robust under simulated reconstruction artifacts (blur, ghosting, and bounding-box noise). 
\end{itemize}

\section{Related Work}
\label{sec:rel_work}
\subsection{Embodied Agents with Vision-Language Models}
\label{subsec:vlm_agents}
Embodied agents aim to perceive, reason, and act in interactive 3D environments. Early embodied systems primarily relied on modular pipelines with task-specific perception, mapping, and planning components, particularly for navigation and object search tasks \cite{anderson2018vision, gupta2017cognitive, chaplot2020object}. While effective in constrained settings, these approaches struggled to generalize to open-ended tasks due to rigid symbolic representations and limited semantic understanding.

Recent advances in large language models (LLMs) and vision--language models (VLMs) have substantially reshaped embodied intelligence. Pretrained multimodal models such as CLIP \cite{radford2021learning}, BLIP \cite{li2022blip}, BLIP-2 \cite{li2023blip}, Flamingo \cite{alayrac2022flamingo}, and LLaVA \cite{liu2023visual} demonstrate strong zero-shot generalization across diverse visual and linguistic tasks. These models enable embodied agents to interpret high-level instructions, reason over observations, and generalize across environments without task-specific retraining. 

Building on these capabilities, recent works on embodied agents leverage language models for action planning and decision-making grounded in visual observations. LEO \cite{huang2023embodied} introduces a multimodal embodied agent that integrates 3D perception with language models to perform diverse tasks such as grounding, reasoning, and navigation in 3D environments. STEVE \cite{zhao2024see} combines visual perception modules with LLM-based reasoning to generate action plans for complex tasks. Octopus \cite{yang2024octopus} formulates embodied control as program generation, where a vision–language model produces executable action code and improves behavior through environmental feedback.

Progress in this direction is driven by embodied navigation and question answering benchmarks. Benchmarks such as Embodied Question Answering (EQA) \cite{das2018embodied}, ScanQA \cite{azuma2022scanqa}, SQA3D \cite{ma2022sqa3d}, OpenEQA \cite{majumdar2024openeqa}, and EXPRESS-Bench \cite{jiang2025beyond} evaluate an agent’s ability to integrate perception, memory, and reasoning over long interaction horizons. Exploration-centric benchmarks such as Explore-EQA \cite{ren2024explore} and GOAT-Bench \cite{khanna2024goat} further emphasize long-horizon cognition and memory utilization. However, enabling such long-horizon, grounded reasoning requires a spatial memory system that can reliably store, organize, and retrieve perceptual information.

\subsection{Spatial Memory Systems for Embodied Agents}
\label{subsec:spatial_memory}

Spatial memory is a central component of embodied reasoning. One line of work adopts view-based memory, where agents retain RGB or RGB-D observations captured during exploration. 3D-Mem \cite{yang20253d} organizes multi-view observations into memory snapshots that capture co-visible objects and scene context for VLM-based reasoning. LMEE \cite{wang2026explore} stores observed images and object metadata that support exploration and question answering. However, reliance on previously observed viewpoints can fail when queries require reasoning from novel or object-centric perspectives.

Object-centric memory representations provide more structured abstractions. Scene graph–based methods \cite{armeni20193d, rosinol20203d} represent environments as objects connected by spatial relations, supporting relational reasoning and planning. Hierarchical and open-vocabulary extensions such as HOV-SG \cite{werby2024hierarchical} and ConceptGraphs \cite{gu2024conceptgraphs} further improve scalability and expressiveness. Real-time spatial perception systems such as Hydra \cite{hughes2022hydra} enable online construction and optimization of such graphs. Nevertheless, graph abstractions often quantize geometry into coarse relations or sparse descriptors, making it difficult to answer viewpoint-dependent visibility queries without explicitly modeling rendering or ray-based reasoning.

Another line of work incorporates semantic representations from pretrained VLMs into 3D scene representations. Methods such as VLMaps \cite{huang2022visual}, OpenScene \cite{peng2023openscene}, ConceptFusion \cite{jatavallabhula2023conceptfusion}, and CLIP-Fields \cite{zhang2023clip} lift language-aligned visual features into 3D maps or neural scene representations, enabling open-vocabulary querying. However, because most vision--language models used by embodied agents operate on 2D image inputs, directly leveraging these 3D representations for reasoning remains challenging. To address this limitation, recent work injects 3D features directly into language models, including 3D-LLM \cite{hong20233d}, LL3DA \cite{chen2024ll3da}, Chat-Scene \cite{huang2024chat}, SplatTalk \cite{thai2025splattalk}, and 3DLLM-Mem \cite{hu20253dllm}. While these approaches improve semantic grounding, they are constrained by context length and computational cost, often requiring aggressive subsampling or pooling that discards fine-grained geometric details. In addition, these representations typically aggregate features in a direction-agnostic manner, making them ill-suited for relational queries that depend on specific viewpoints.

Overall, while prior spatial memory systems have made substantial progress in long-horizon reasoning and semantic grounding, they remain limited in supporting viewpoint-dependent visibility. This limitation arises from a mismatch between how scenes are represented in memory and how embodied queries specify perspectives. RenderMem addresses this gap by treating rendering itself as a memory operation, generating query-conditioned visual evidence from object-centric viewpoints that can be directly consumed by vision-language models.

\section{Method}
\label{sec:method}
RenderMem is built on a simple but powerful abstraction: rendering is the read operation of spatial memory. Rather than treating rendering as a visualization tool, we elevate it to a first-class mechanism for querying geometry. This design decouples spatial reasoning from language reasoning while preserving compatibility with existing vision--language models.

Instead of storing fixed observations or converting raw 3D representations into tokens, RenderMem maintains a persistent, renderable 3D scene state and generates visual evidence only when required by a query. In this framework, a question does not retrieve stored images--it triggers viewpoint-conditioned rendering that computes the exact visual evidence necessary for reasoning about visibility and occlusion.

\begin{figure}[tbp]
    \centering 
    \includegraphics[width=\columnwidth]{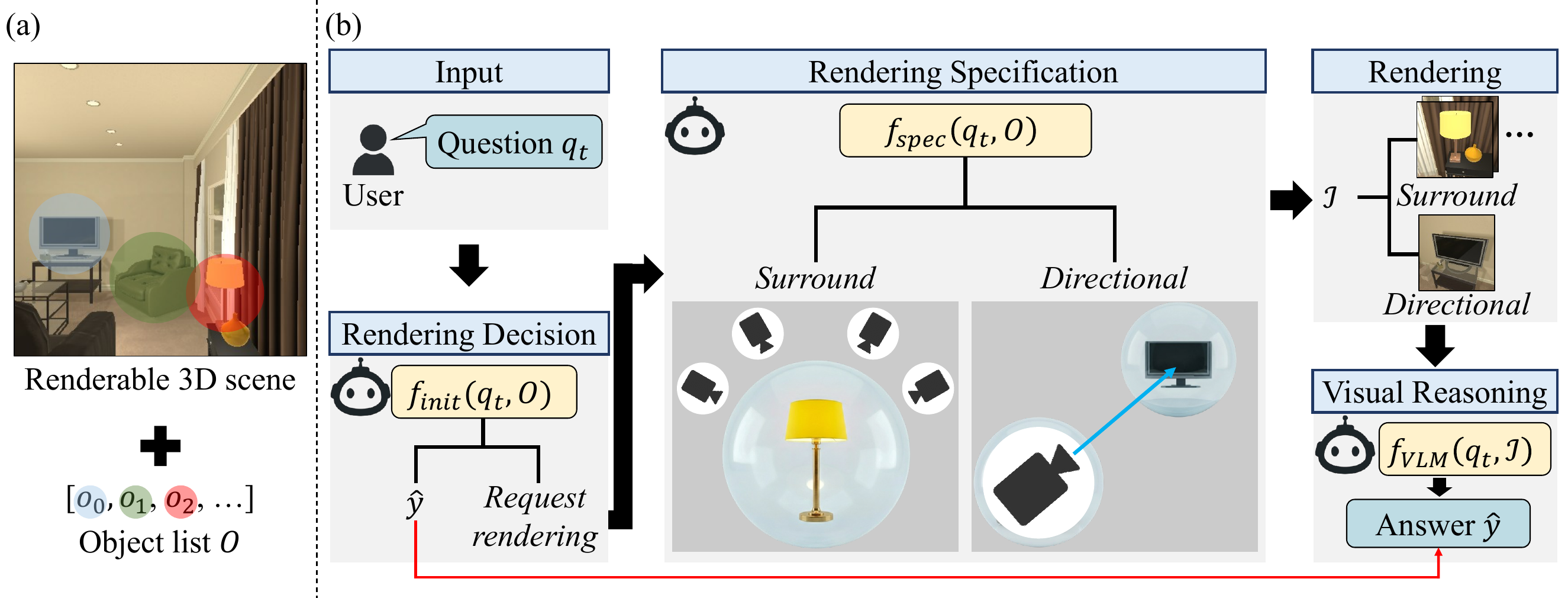} 
    \caption{Overview of the RenderMem pipeline. (a) A renderable 3D scene with an object list serves as spatial memory. (b) Given a question, RenderMem decides whether rendering is needed and selects a rendering mode and object anchors. Surround rendering captures multiple views around an object, while directional rendering generates a source-to-target viewpoint for visibility reasoning. The rendered images are used to answer the user question.}
    \label{fig:pipeline}
\end{figure}

Given a question, RenderMem executes a structured two-stage pipeline that determines (1) whether rendering is required, (2) what type of rendering method is most appropriate and which objects should guide the rendering. The rendered images are then provided to a vision--language model together with the original question to produce the final answer. The pipeline is visualized in~\cref{fig:pipeline}.

\subsection{Scene Representation}
\label{subsec:scene_rep}
RenderMem assumes access to a renderable 3D scene representation that serves as a persistent and unified spatial memory of the environment.
In practical deployments, such scene representations can be incrementally constructed and updated using established mapping pipelines.
Camera trajectories may be recovered via SLAM~\cite{kim2018linear, lamarca2018camera}, while object instances are detected, localized, and tracked over time. 
The underlying geometry can be maintained in a renderable form using mesh reconstruction~\cite{kanazawa2018learning}, neural radiance fields~\cite{mildenhall2021nerf}, or 3D Gaussian Splatting~\cite{kerbl20233d}. 
As new observations arrive, the scene estimate is progressively refined to reflect the current state of the environment.

Importantly, RenderMem is agnostic to the specific reconstruction backend and assumes only access to the current scene estimate. 
Rather than storing historical observations, the system retrieves spatial evidence through query-conditioned rendering of the latest scene state. 
Because rendering operates directly on the current geometry, spatial evidence is always derived from the most up-to-date map. 
Consequently, when the underlying scene representation evolves due to environmental changes, RenderMem naturally adapts without requiring explicit memory updates. 

Given such a renderable scene state, we introduce a lightweight object-level abstraction that supports query-conditioned viewpoint specification. This abstraction provides stable geometric anchors for camera placement while avoiding direct exposure of raw 3D geometry to the language model. Formally, we represent the scene as a set of $N$ objects
\begin{equation}
\mathcal{O} = \{ o_i \}_{i=1}^N
\label{eq:object_list}
\end{equation}
where each object $o_i$ is represented as
\begin{equation}
o_i = (\text{id}_i,\; \mathbf{s}_i)
\label{eq:object_def}
\end{equation}
Here, $\text{id}_i$ denotes a unique identifier composed of an object category and an index (e.g., Chair\_0), and $\mathbf{s}_i$ denotes a bounding sphere encoding the object's spatial extent.

To balance geometric expressiveness and efficiency, each object is approximated by a bounding sphere derived from its axis-aligned bounding box. Given the eight bounding-box corner points $\{ \mathbf{p}_{ij} \}_{j=1}^8 \subset \mathbb{R}^3$, the object center and radius are computed as
\begin{align}
\mathbf{c}_i &= \frac{1}{8} \sum_{j=1}^8 \mathbf{p}_{ij}, \label{eq:center} \\
r_i &= \max_{1 \le j \le 8} \| \mathbf{p}_{ij} - \mathbf{c}_i \|_2.
\label{eq:radius}
\end{align}
This spherical approximation captures the object’s position and scale while remaining invariant to orientation and topology. 

\subsection{Question-Answering Pipeline}

RenderMem answers each user question $q_t$ posed over the current scene by issuing a sequence of internal queries that determine how visual evidence should be retrieved. While the question represents the task posed over the scene, the queries correspond to intermediate decisions that control the rendering process. This procedure consists of two stages: (1) determining whether rendering is necessary, and (2) if so, producing a query-conditioned rendering specification. The final answer is then obtained by applying a vision–language model to the rendered views.

\paragraph{Query 1: Rendering Decision.}
Given a question $q_t$ and the current object list $\mathcal{O}$, RenderMem first issues an internal query to determine whether explicit visual evidence is required or whether the question can be answered directly from $\mathcal{O}$:
\begin{equation}
f_{\text{init}}(q_t, \mathcal{O}) =
\begin{cases}
\hat{y}, & \text{if answerable}, \\
\texttt{request\_rendering}, & \text{otherwise}.
\end{cases}
\label{eq:init}
\end{equation}
where $\hat{y}$ denotes the predicted answer. This gating step avoids unnecessary rendering for queries that can be answered directly from the object list, such as counting object instances (e.g., ``How many chairs are in the room?'').

\paragraph{Query 2: Rendering Specification.}
If rendering is requested, RenderMem issues a second internal query that produces a structured rendering specification. This specification jointly determines the rendering mode and the object anchor(s) used to guide camera placement:
\begin{equation}
f_{\text{spec}}(q_t, \mathcal{O}) = \rho = (m, \mathcal{A}),
\label{eq:spec}
\end{equation}
where $m \in \{\text{surround}, \text{directional}\}$ and $\mathcal{A}$ denotes the mode-dependent object anchors:
\begin{equation}
\mathcal{A} =
\begin{cases}
\{o_i\}, & m = \text{surround}, \\
(o_s, o_t), & m = \text{directional}.
\end{cases}
\label{eq:anchors}
\end{equation}
The surround mode captures multiple views around a single object and is suitable for queries that require observing object attributes or states. The directional mode supports viewpoint-dependent reasoning such as visibility and occlusion, where the viewpoint is instantiated relative to a source object $o_s$ and a target object $o_t$.
By specifying object anchors rather than raw geometry, the rendering process focuses on spatial regions relevant to $q_t$ while introducing a structured abstraction that is more amenable to language-based reasoning. This abstraction enables the system to translate symbolic queries into geometrically grounded renderings while maintaining a compact interface between the 3D scene representation and the language model. Algorithmic details are provided in \cref{subsec:rendering_algorithms}.

\paragraph{Scene Rendering.}
Given the rendering specification $\rho$, the renderer produces a set of images
\begin{equation}
\mathcal{I} = \textsc{Render}(\mathcal{S}, \mathcal{O}, \rho),
\label{eq:render}
\end{equation}
where $\mathcal{S}$ denotes the underlying renderable scene state.
The rendering process instantiates camera viewpoints relative to the selected object anchors and produces views that capture spatial relationships relevant to the question.

\paragraph{Evidence-Based Reasoning.}
 The rendered images $\mathcal{I}$ serve as query-conditioned visual evidence grounded in the current scene geometry. These images are paired with the original question and passed to a vision–language model to produce the final answer:
\begin{equation}
\hat{y} = f_{\text{VLM}}(q_t, \mathcal{I}).
\label{eq:vlm}
\end{equation}

By transforming symbolic questions into rendered observations, RenderMem enables the vision--language model to reason over explicit visual evidence grounded in the scene geometry.

\subsection{Rendering Algorithms}
\label{subsec:rendering_algorithms}

We formulate rendering as the problem of selecting camera parameters: a camera position $\mathbf{x} \in \mathbb{R}^3$ and a viewing direction $\mathbf{v} \in \mathbb{R}^3$. RenderMem uses camera placement as a mechanism for producing query-conditioned observations that support grounded reasoning about the scene.

Under this formulation, different rendering strategies correspond to different constraints on how the camera parameters $(\mathbf{x}, \mathbf{v})$ are chosen. Intuitively, the goal is not merely to capture the scene but to synthesize views that expose the spatial relationships required to answer the question. RenderMem instantiates this idea using two complementary rendering modes: \emph{surround rendering}, which provides contextual observations around a target object, and \emph{directional rendering}, which simulates viewpoint-specific observations between objects for visibility and occlusion reasoning. 

\paragraph{Surround Rendering.}
Surround rendering generates multiple views around a target object to provide sufficient visual evidence about its appearance and surrounding context. Given the object sphere $\{ (\mathbf{c}_i, r_i) \}$, we compute an enclosing camera sphere $(\mathbf{c}, R_{\text{cam}})$, where $\mathbf{c}$ denotes the object center and $R_{\text{cam}}$ represents the minimum radius required to place cameras outside the object geometry.

To ensure that the entire object remains fully visible in the rendered views, we first compute the minimum camera distance required for the object sphere to fit within the view frustum. Let $\mathrm{FOV}_v$ denote the vertical field of view and $\mathrm{aspect}$ the image aspect ratio. Since both vertical and horizontal field-of-view constraints must be satisfied, we define an effective half field-of-view angle
\begin{equation}
\beta =
\min\!\left(
\frac{\mathrm{FOV}_v}{2},
\arctan\!\left(
\tan\!\left(\frac{\mathrm{FOV}_v}{2}\right)\cdot \mathrm{aspect}
\right)
\right).
\label{eq:fov}
\end{equation}
Using basic trigonometric relations, the minimum camera distance from the object center that guarantees full object visibility is
\begin{equation}
d_{\min} = \frac{r_i}{\sin(\beta)}.
\label{eq:dmin}
\end{equation}

To control the amount of contextual information included in the rendered images, we scale this minimum distance with a factor $\alpha$
\begin{equation}
d = \alpha\, d_{\min}.
\label{eq:distance}
\end{equation}
Larger values of $\alpha$ increase the viewing distance and reveal more surrounding context, while smaller values produce tighter object-centric views.

In practice, single viewpoints may fail to capture relevant evidence due to occlusions caused by nearby objects. To improve robustness, RenderMem samples $K$ camera poses uniformly around the object along the azimuth direction while maintaining a fixed elevation angle $\phi$. This strategy ensures that at least some views expose the target object without occlusion while still preserving consistent spatial context across views.
For azimuth angles $\theta_i = \frac{2\pi i}{K}$, the viewing direction is
\begin{equation}
\mathbf{v}_i =
\mathrm{normalize}\!\left(
(\mathbf{p}\cos\theta_i + \mathbf{q}\sin\theta_i)\cos\phi
+ \mathbf{u}\sin\phi
\right).
\label{eq:viewdir}
\end{equation}
where $\mathbf{u}$ denotes the world up vector and $\mathbf{p}, \mathbf{q}$ form an orthonormal basis spanning the horizontal plane. Each camera is placed at
\begin{equation}
\mathbf{x}_i = \mathbf{c} + d\,\mathbf{v}_i.
\label{eq:campos}
\end{equation}
and oriented to look toward the object center $\mathbf{c}$.

\paragraph{Directional Rendering.}
Directional rendering uses a single camera pose to reason about visibility and occlusion from a source object toward a target object. Given the source sphere $(\mathbf{c}_s, r_s)$ and the target center $\mathbf{c}_t$, the camera is placed on the surface of the source sphere at the point closest to the target to avoid self-occlusion by the source object, and oriented toward the target center. This configuration approximates the viewpoint from the source object toward the target, enabling reasoning about visibility and occlusion between the two objects.

\section{Experiments}
\label{sec:experiments}

\subsection{Benchmark}
\label{subsec:experiments}
We build our benchmark on indoor environments from AI2-THOR\cite{kolve2017ai2}, including iTHOR\cite{kolve2017ai2}, RoboTHOR\cite{deitke2020robothor}, and ProcTHOR\cite{procthor} scenes. The benchmark is designed to evaluate different levels of spatial reasoning required for embodied agents. 
The benchmark consists of three subsets: 1) \textit{Static Attribute and Count} subset evaluates object-centric reasoning in static scenes, 2) \textit{Dynamic Attribute and Count} subset evaluates reasoning over state changes caused by interactions, and 3) \textit{Viewpoint-Dependent Visibility} subset evaluates whether a target object is visible from a specified spatial location, requiring viewpoint-conditioned geometric reasoning.
The benchmark spans 180 scenes across multiple AI2-THOR environments. Benchmark details are provided in the supplementary.

\subsection{Binary LLM Match Evaluation}
\label{subsec:binary_llm_match}
Evaluating open-ended question answering with language models presents a challenge: although ground-truth answers are represented in canonical forms (e.g., \textit{yes}/\textit{no} or numeric counts), model predictions are often expressed in unconstrained natural language. For instance, a model may answer ``closed" instead of ``yes”, or ``there are two chairs” instead of ``2”. While semantically correct, such responses would be marked incorrect under exact string matching.

To address this issue, we adopt a binary variant of the \textit{LLM Match} metric~\cite{majumdar2024openeqa}. Instead of strict string comparison, an external evaluation LLM determines whether a predicted answer $\hat{y}$ correctly answers the original question given the ground-truth answer $y$:
\begin{equation}
\mathrm{Match}(\hat{y}, y) =
\begin{cases}
1, & \text{if } \hat{y} \text{ is semantically identical to $y$}, \\
0, & \text{otherwise}.
\end{cases}
\label{eq:binary_match}
\end{equation}

\subsection{Baselines}

We compare RenderMem against three representative classes of baselines: \textit{Multi-view retrieval}~\cite{wu2019unified, radford2021learning}, \textit{Concept Graphs}~\cite{gu2024conceptgraphs}, and \textit{3D-Mem}~\cite{yang20253d}. \textit{Multi-view retrieval} stores all explored views and retrieves the most relevant images for each question using an image–text similarity model~\cite{radford2021learning}. \textit{Concept Graphs} represent the scene as an object-centric graph with textual descriptions of object instances and perform reasoning directly over these descriptions. \textit{3D-Mem} maintains a hybrid memory consisting of object lists and view-based snapshots that associate images with object instances and spatial attributes. Implementation details are provided in the supplementary.




\subsection{Comparative Analysis}
\label{subsec:comparative}
We compare RenderMem with representative spatial memory approaches to understand how different memory representations affect spatial question answering. All methods are evaluated on the same QA sets, and answers are generated using the same Qwen2.5-VL-7B \cite{qwen2.5-VL} model to ensure a fair comparison. Performance is reported using the averaged binary LLM Match score. Results are reported in~\cref{tab:main_results}.

\paragraph{Object QA}
This task evaluates reasoning over object attributes (e.g., ``Is the TV turned on?'') and object counts (e.g., ``How many chairs are in the room?'') in a static environment. 


\begin{table}[t]
\centering
\caption{Comparison across spatial reasoning tasks.}
\label{tab:main_results}
\setlength{\tabcolsep}{6pt}
\begin{tabular}{lccccc}
\toprule
& \multicolumn{2}{c}{Object QA} & \multicolumn{3}{c}{Visibility QA} \\
\cmidrule(lr){2-3} \cmidrule(lr){4-6}
Method & Attribute & Count & RoboTHOR & ProcTHOR & Avg \\
\midrule
Multi-view retrieval & 0.69 & 0.25 & 0.50 & 0.49 & 0.50 \\
Concept Graphs & 0.60 & 0.53 & -- & -- & -- \\
3D-Mem & 0.68 & 0.78 & 0.42 & 0.43 & 0.43 \\
\textbf{RenderMem (ours)} & \textbf{0.82} & \textbf{0.78} & \textbf{0.81} & \textbf{0.77} & \textbf{0.79} \\
\bottomrule
\end{tabular}
\end{table}

Multi-view retrieval performs reasonably on attribute questions by leveraging visual cues, but struggles on counting since raw images do not provide explicit instance aggregation. Concept Graphs improve counting through object-level nodes, yet performance on attribute questions suffers when relevant visual details are not preserved in the generated descriptions. 3D-Mem achieves strong counting via explicit instance tracking, yet attribute accuracy remains limited because the snapshots are optimized to capture many objects per view, often diluting object-specific visual cues. 

RenderMem achieves the best overall performance by generating views conditioned on the query, preserving both object completeness and detailed visual cues. 

\paragraph{Visibility QA}
We further evaluate viewpoint-conditioned visibility reasoning (e.g., ``Is the TV visible from the sofa?'').
Multi-view retrieval achieves moderate performance because retrieved views often contain both queried objects but are not aligned with the requested viewpoint. 3D-Mem suffers from limited viewpoint coverage since it preserves only a compact subset of snapshots, reducing the likelihood of retrieving a view consistent with the queried spatial configuration.

In contrast, RenderMem explicitly renders views conditioned on the queried spatial relation. By generating a camera pose that corresponds to the specified viewpoint, the method provides geometrically aligned visual evidence for the vision--language model. This significantly reduces ambiguity in visibility reasoning and leads to large performance gains across both RoboTHOR and ProcTHOR scenes.

\subsection{Support for Dynamic Scenes}
\label{subsec:dynamic_mem}
Beyond static scenes, we analyze how RenderMem behaves in dynamic environments where object states change due to interactions. Unlike conventional memory systems that explicitly maintain and update scene states, RenderMem generates visual evidence at query time from the current scene representation. As a result, changes in object appearance are immediately reflected in the rendered observations without requiring explicit memory updates.

We evaluate this property using the \textit{Dynamic Attribute and Count} dataset, which consists of interaction–question pairs where each interaction modifies the scene state before the question is asked (e.g., turning on a TV followed by ``Is the TV turned on?''). The interaction types follow the affordances provided by AI2-THOR, including clean, dirty, break, slice, and toggle actions.

Our analysis reveals that RenderMem performs robustly in dynamic environments where object states change due to interactions. In dynamic scenes, RenderMem achieves 0.92 attribute accuracy and 0.82 counting accuracy, compared to 0.82 and 0.78 in static scenes. Interestingly, performance is slightly higher in the dynamic setting. We speculate that interactions often produce informative state changes, making object attributes more visually distinctive (e.g., toggling a TV or opening a container). Further details are provided in the supplementary.

\subsection{Multi-Step Querying}
\label{subsec:multi_step_query}
We investigate how the decision process for rendering should be structured. A single query must determine three factors simultaneously: whether rendering is required, how to render (mode), and what to render (object anchors). Solving these decisions jointly can impose a substantial reasoning burden on the language model.

To study the effect of structured decomposition, we evaluate three strategies that progressively separate these decisions. The \textit{1-step} setting predicts rendering necessity, rendering mode, and object anchors in a single query. The \textit{2-step} setting first determines whether rendering is required, and if so, jointly predicts rendering mode and object anchors. The \textit{3-step} setting resolves rendering necessity, rendering mode, and object anchors through three sequential queries.

Across all tasks, all strategies show comparable LLM Match score. The 2-step formulation performs best (0.80 / 0.87 / 0.79 on static, dynamic, and visibility tasks), improving over the 1-step formulation (0.78 / 0.85 / 0.78) while maintaining the coupling between rendering mode and object anchors. The 3-step formulation slightly degrades visibility performance (0.80 / 0.86 / 0.77), suggesting that fully separating these decisions can break dependencies that jointly determine the camera pose.
We therefore adopt the two-step formulation in the final system.

\begin{figure}[tbp]
    \centering 
    \includegraphics[width=\columnwidth]{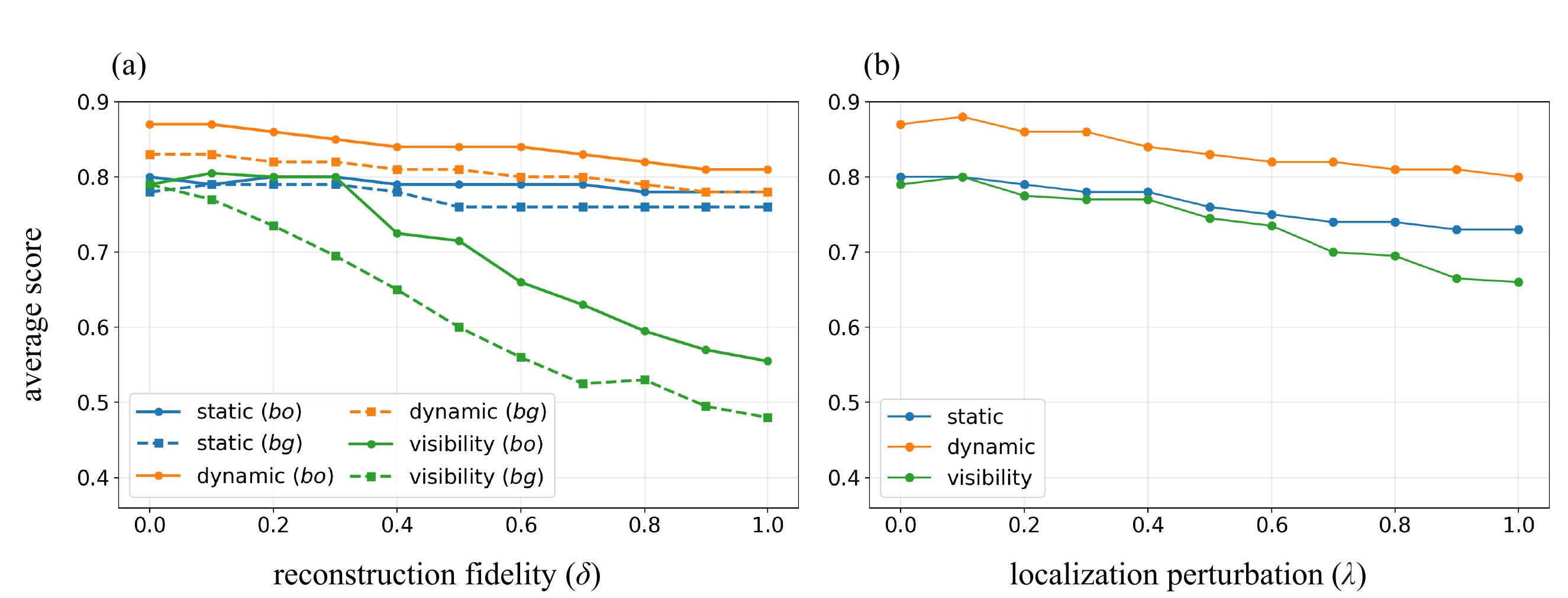} 
    \caption{RenderMem's robustness to imperfect scene representations. (a) Performance under decreasing reconstruction fidelity, simulated with blur-only (\textit{bo}) and blur+ghosting (\textit{bg}). (b) Performance under increasing localization perturbation applied to object bounding spheres.}
    \label{fig:robustness}
\end{figure}

\subsection{Robustness to Imperfect Scene Representations}
\label{subsec:robustness}
Real-world scene representations are rarely perfect due to sensor noise, incomplete reconstruction, or localization errors. To assess practical robustness, we evaluate RenderMem on object QA under both static and dynamic conditions, and on viewpoint-dependent visibility QA in static scenes, as visibility reasoning is independent of scene interactions. We simulate two common imperfections: degraded reconstruction fidelity and inaccurate object localization.

\paragraph{Reconstruction Fidelity.}
We first study robustness to low-quality scene reconstructions by corrupting rendered images $\mathcal{I}$ with controlled image-level artifacts that mimic common reconstruction errors: blur and ghosting.

Blur is applied via Gaussian smoothing:
\begin{equation}
\mathcal{I}' = \mathrm{GaussianBlur}(\mathcal{I}, \sigma),
\quad
\sigma = 0.5 + 6\delta,
\end{equation}
where $\delta \ge 0$ controls the reconstruction degradation severity.

Ghosting is simulated by blending spatially shifted image copies:
\begin{equation}
\mathcal{I}' = (1-\gamma)\mathcal{I}
+ 0.6\gamma\,\mathcal{T}_{\Delta_1}(\mathcal{I})
+ 0.4\gamma\,\mathcal{T}_{\Delta_2}(\mathcal{I}),
\end{equation}
where $\mathcal{T}_{\Delta}$ denotes translation by a pixel offset $\Delta$, and $\gamma \in [0,1]$ controls the ghosting intensity.

\Cref{fig:robustness}(a) shows that object QA performance for static and dynamic remains stable across a wide range of reconstruction degradation under both blur-only (\textit{bo}) and blur$+$ghosting (\textit{bg}) settings. This robustness stems from RenderMem's object-centric close-up views, which preserve discriminative cues despite reconstruction artifacts.
Visibility reasoning is more sensitive to reconstruction quality because distant objects occupy smaller image regions, where blur or ghosting can obscure spatial details and complicate occlusion reasoning.

\paragraph{Localization Perturbation.}
We next evaluate robustness to imperfect object localization by injecting geometric noise into the object bounding spheres $\mathbf{s}_i = (\mathbf{c}_i, r_i)$ defined in \cref{subsec:scene_rep}. 

The center is perturbed as
\begin{equation}
\tilde{\mathbf{c}}_i = \mathbf{c}_i + \boldsymbol{\epsilon}, 
\quad 
\boldsymbol{\epsilon} \sim \mathcal{N}(\mathbf{0}, (\lambda r_i)^2 \mathbf{I}),
\end{equation}

and the radius as
\begin{equation}
\tilde{r}_i = r_i (1 + \xi), 
\quad 
\xi \sim \mathcal{N}(0, \lambda^2).
\end{equation}

where $\lambda \ge 0$ controls the magnitude of the localization perturbation.

\Cref{fig:robustness}(b) shows that object QA performance for static and dynamic remains largely stable under increasing localization noise, indicating that moderate geometric inaccuracies do not immediately invalidate rendered evidence as long as target objects remain visible. Visibility queries again exhibit higher sensitivity due to their reliance on precise viewpoint geometry and occlusion relationships. Nevertheless, performance remains robust under realistic levels of localization perturbation, demonstrating that RenderMem tolerates practical scene representation errors.


\section{Limitations and Future Work}
\label{sec:limitations}
Despite its effectiveness, RenderMem has several limitations.

First, the current object abstraction relies primarily on object category identifiers and spatial locations. This representation lacks support for fine-grained instance-level disambiguation when multiple objects of the same category are present within a scene. For example, when a question refers to ``the chair next to the window'' in a room containing multiple chairs, the system struggles to disambiguate the intended instance. Augmenting each object with visual features extracted from multi-view observations could potentially address this issue. However, such an approach would introduce significant computational overhead and would complicate support for dynamic environments, as visual features would need to be recomputed after every scene interaction. A more desirable direction is the development of a lightweight mechanism that captures an object’s surrounding spatial context without requiring repeated visual feature extraction, thereby preserving computational efficiency while maintaining compatibility with dynamic scene updates.

Second, RenderMem relies on a renderable 3D scene representation whose quality directly affects the fidelity of the rendered evidence used for reasoning. High-quality scene reconstructions typically require substantial storage. For example, indoor mesh scenes can occupy approximately 850\,MB per scene, while hybrid mesh--3DGS representations may require around 220\,MB~\cite{miao2025towards}. When many environments must be stored, such requirements can become a practical storage bottleneck. Exploring more storage-efficient scene representations or compression strategies that preserve geometric fidelity could improve the scalability of RenderMem to large numbers of environments.

\section{Conclusion}
\label{sec:conclusion}

We present RenderMem, a spatial memory framework that leverages rendering as the read operation of a 3D scene representation. Instead of retrieving stored observations, RenderMem generates query-conditioned visual evidence by rendering the scene from viewpoints relevant to the user question. This design explicitly grounds reasoning in geometry while remaining fully compatible with existing vision–language models. 

Experiments demonstrate that this approach improves spatial question answering, particularly for viewpoint-dependent visibility reasoning where prior memory paradigms struggle. Because visual evidence is synthesized directly from the current scene state, RenderMem naturally supports dynamic environments without requiring explicit memory updates. We hope this work motivates further research on rendering-based spatial memory systems and inspires geometry-aware architectures that bridge 3D scene representations and language-based reasoning for embodied intelligence.

\bibliographystyle{splncs04}
\bibliography{main}
\end{document}